\documentclass{article}
\usepackage{spconf,amsmath,graphicx}
\usepackage{threeparttable}
\usepackage{cite}
\usepackage{setspace}
\usepackage[colorlinks,linkcolor=blue]{hyperref}

\begin{document}
\title{Multi-Head Feature Pyramid Networks for Breast Mass Detection}

\name{\begin{tabular}{c}Hexiang Zhang$^{1}$, Zhenghua Xu$^{1,\dag,\star}$\thanks{$^{\star}$Co-first author.}, Dan Yao$^{1}$, Shuo Zhang$^{1}$, Junyang Chen$^{2}$, Thomas Lukasiewicz$^{3,4}$\thanks{$^{\dag}$Corresponding authors: zhenghua.xu@hebut.edu.cn (Zhenghua Xu)}\end{tabular}}
\address{
$^1$State Key Laboratory of Reliability and Intelligence of Electrical Equipment, \\
School of Health Sciences and Biomedical Engineering, Hebei University of Technology, China\\
$^2$College of Computer Science and Software Engineering and Guangdong Laboratory
of Artificial\\ Intelligence and Digital Economy (SZ), Shenzhen University, Shenzhen, China\\
$^3$Institute of Logic and Computation, TU Wien, Vienna, Austria\\
$^4$Department of Computer Science, University of Oxford, United Kingdom
}

\maketitle
\begin{abstract}
Analysis of X-ray images is one of the main tools to diagnose breast cancer. The ability to quickly and accurately detect the location of masses from the huge amount of image data is the key to reducing the morbidity and mortality of breast cancer. Currently, the main factor limiting the accuracy of breast mass detection is the unequal focus on the mass boxes, leading the network to focus too much on larger masses at the expense of smaller ones. In the paper, we propose the multi-head feature pyramid module (MHFPN) to solve the problem of unbalanced focus of target boxes during feature map fusion and design a multi-head breast mass detection network (MBMDnet). Experimental studies show that, comparing to the SOTA detection baselines, our method improves by 6.58\% (in AP@50) and 5.4\% (in TPR@50) on the commonly used INbreast dataset, while about 6-8\% improvements (in AP@20) are also observed on the public MIAS and BCS-DBT datasets. 
\end{abstract}

\vspace*{-2.0ex}
\section{INTRODUCTION}
\vspace*{-1.5ex}
\label{sec:intro}
Breast cancer has become a major disease threatening women's health in terms of morbidity and mortality. According to the statistics~\cite{bray2018global}, in 2018 alone, the number of new cases reached 2,088,849 (11.6\%) and deaths were 626,679 (6.6\%). By comparing the morbidity and mortality rates of breast cancer, it is clear that timely diagnosis and treatment at an early stage can significantly reduce the mortality rate~\cite{yu2022systematic}. Currently, the detection of breast masses is mainly divided into: ultrasound image detection, X-ray image detection and biopsy. Among them, X-ray has become the main modality for the early detection of breast cancer due to its outstanding imaging performance. However, X-ray images require specialized radiologists to screen a large number of Digital Imaging and Communications in Medicine (DICOM) files, and the process of resolving X-ray images is very tedious and time-consuming~\cite{gastounioti2022artificial}. In order to improve the accuracy and efficiency of breast mass detection, many computer-aided detection methods based on deep learning have been rapidly developed and achieved good results~\cite{zhang2022anchor,ibrokhimov2022two,su2022yolo,PAC-net}.

\begin{small}
\begin{figure*}[htp]
\centering
\vspace{-2.0em}
\includegraphics[width=180mm,height=66.47mm]{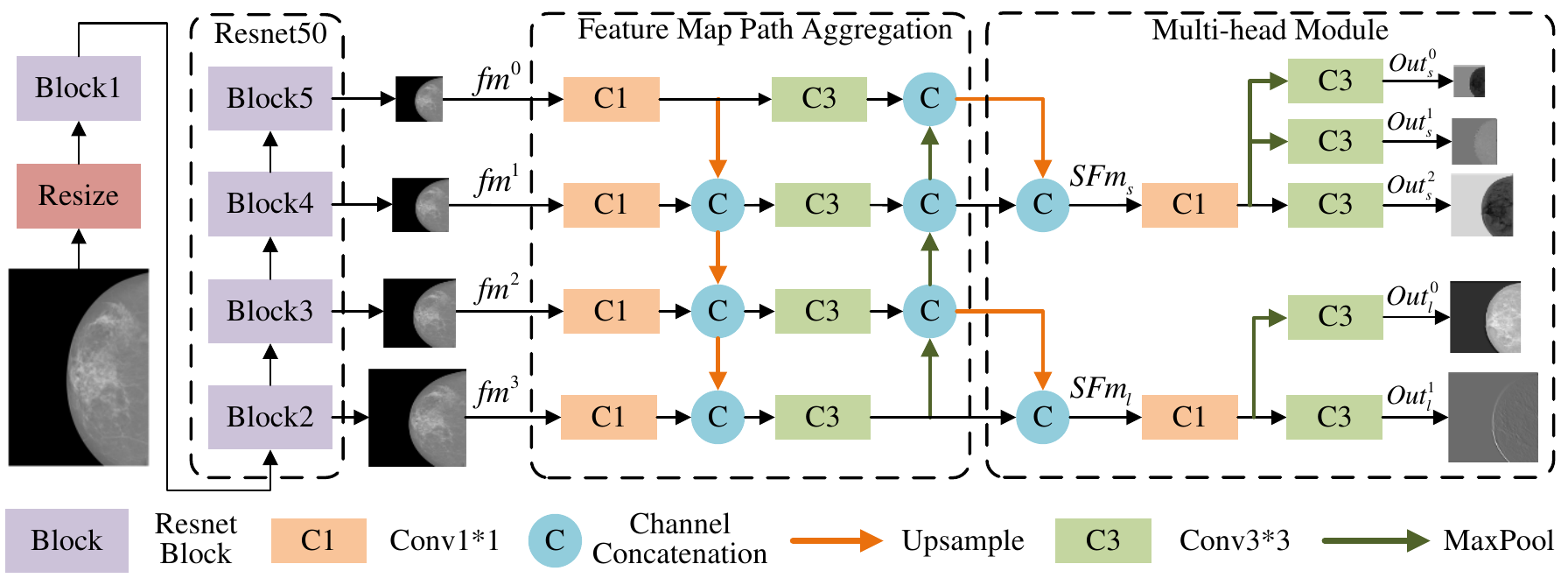}
\caption{Overview of multi-head high resolution FPN network. Since the head part of the detection network follows the original structure, the head part has been omitted from the overall structure diagram.}
\vspace{-1.5em}
\label{fig:MHFPN}
\end{figure*}
\end{small}

Most of the breast mass detection studies are based on the classical network to adapt the network structure to the characteristics of X-ray images. L. Zhang et al.~\cite{zhang2022anchor} proposed an Anchor-free YOLOv3 network to alleviate the imbalance between positive boxes and negative boxes that occurs in breast mass detection. B. Ibrokhimov et al.~\cite{ibrokhimov2022two} proposed a slider-cutting image method based on Faster RCNN to generate small square patches to solve the problem of excessive X-ray image size and resolution. Y. Wu et al.~\cite{wu2018automatic} first proposed the use of the Faster RCNN network based on hierarchical candidate frames for breast mass detection. The approach not only reduces the amount of data to be processed but also improves the accuracy of detection. H. Cao et al.~\cite{cao2021breast} then proposed a new normalization approach and image enhancement algorithm to further improve the detection accuracy of breast masses based on the FSAF network structure. The aforementioned methods have achieved excellent performance on public datasets such as INbreast~\cite{moreira2012inbreast} or private datasets; however, they have a common shortcoming in that they focus more on the features of large masses and ignore those of small masses during feature fusion, which thus leads to serious medical misdiagnosis and may endanger the health of patients in clinical practices.

To solve the unequal focus in the feature fusion of breast X-ray images in target detection, We propose a parallel multi-head feature pyramid module (MHFPN). By adding two additional aggregated focus heads, the network can more easily focus on breast masses that make up a smaller percentage of the image, resulting in improved detection accuracy. Based on MHFPN, Faster RCNN is selected as the detection network. By testing on three publicly available breast mass datasets, our method can achieve the same performance as the previously mentioned X-ray breast image detection studies with a simpler network structure.

In general, our contributions are three-folded: (1) We design a general MHFPN on the feature map fusion layer to alleviate the problem of unequal focus in the feature fusion stage. (2) We propose a multi-headed breast mass detection network (MBMDnet) to improve the accuracy of mass target detection and reduce the miss detection rate. (3) Experiments show that our method improves by 6.58\% (in AP@50) and 5.4\% (in TPR@50) on the commonly used INbreast dataset, and 6\%-8\% (in AP@20) on the MIAS and BCS-DBT datasets.
\begin{small}
\begin{figure*}[!t]
\vspace{-2em}
\centering
\includegraphics[width=170mm]{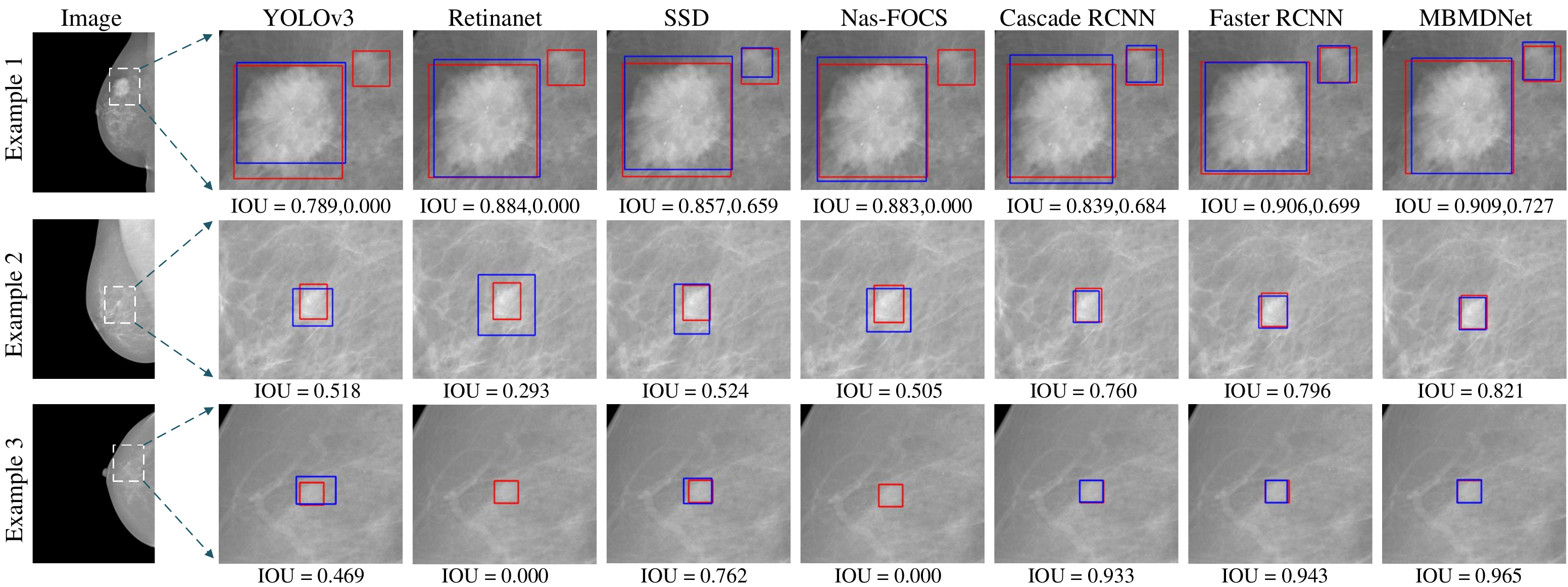}
\vspace{-1em}
\caption{Visualized results of different models on INbreast, and blue (resp., red) boxes are detection results (resp., ground truths).}
\label{fig:MHFPN compare}
\vspace{-1.5em}
\end{figure*}
\end{small}

\begin{small}
\begin{table*}[ht]
\centering
\caption{Results of MBMDNet and classical object detection networks on the INbreast dataset}
\label{table:1}
\vspace{0.3em}
\begin{tabular}{cccccc}
\hline
 Model        & AP@50               & AP@75           & TPR@50         & FLOPS   & Params  \\ \hline
 YOLOv3~\cite{redmon2018yolov3}     & 0.3700 (0.0242)  & 0.1288 (0.0544) & 0.8160 (0.0717) & 19.39G  & 61.53M  \\
 Retinanet~\cite{lin2017focal}      & 0.3774 (0.0652)  & 0.0920 (0.0325) & 0.8226 (0.0823) & 81.89G  & 36.14M  \\
 SSD~\cite{liu2016ssd}              & 0.4892 (0.0889)  & 0.1526 (0.1106) & 0.7590 (0.1604) & 87.86G  & 24.53M \\
 Nas-FCOS~\cite{wang2020fcos}       & 0.4170 (0.1210)  & 0.0872 (0.0367) & 0.8336 (0.0587) & 201.57G & 41.13M \\
 Cascade RCNN~\cite{cai2018cascade} & 0.5212 (0.0509)  & 0.3092 (0.1135) & 0.7354 (0.0867) & 244.11G & 68.93M \\
 Faster RCNN~\cite{ren2015faster}   & 0.5804 (0.0730)  & \textbf{0.3160 (0.1577)} & 0.7716 (0.0633) & 216.30G & 41.13M \\ 
 \textbf{MBMDNet}                            &\textbf{0.6462 (0.0934)} & 0.3034 (0.1035) &\textbf{0.8390 (0.0915)} &291.36G &44.80M \\ \hline
\end{tabular}
\vspace{-1.5em}    
\end{table*}
\end{small}

\vspace{-1em}
\section{METHODS}
\vspace{-1em}

\subsection{Overall Structure}
\vspace{-0.5em}

Inspired by PANet~\cite{liu2018path} and HRFPN~\cite{sun2019high}, we propose a parallel multi-head feature pyramid network to improve the accuracy of breast mass detection in X-Ray images. The network consists of three parts, the backbone for extracting features, the feature map path aggregation, and the multi-head module. The overall structure is shown in Fig~\ref{fig:MHFPN}, where the network is Faster RCNN, and the backbone is selected as Resnet50. At the model input, the image size of the INbreast dataset is $3328 \times 4084$, which is oversized for the existing target detection network breast X-ray images. Due to the limitation of computational volume and operation speed, MHFPN uses resized images ($1333 \times 800$) as the input of the network. Then, the images are fed into the structure of the backbone to obtain feature maps (Denoted as $fm^{i},i\in (0,3)$) at different scales. After that, the feature map is sent through a path aggregation network of successive up and down paths to obtain the richer feature maps (Denoted as $P(fm),f\in (fm^{0},fm^{3})$). 

\vspace{-1em}
\subsection{Multi-head Module}
\vspace{-0.5em}

After the multi-resolution feature maps are aggregated by the feature map paths composed of multiple pooling and convolution layers, the feature information flow is fully fused. Compared with the feature-extracted $fm^{i}$, $P(fm^{i})$ not only carries the precise location information in the bottom information, but also retains the useful information in each feature level. Because of the long path way of information experienced, the X-ray breast feature map suffers from a distraction problem, making the focus of masses with a small image ratio much smaller than that of masses with a large image ratio. Therefore, we designed the multi-head module to alleviate the inequality of focus. Through two feature aggregation headers, the multi-resolution feature maps are assigned focus using the re-convolution to obtain the aggregated feature maps (Denoted as $SFm_{s}$ and $SFm_{l}$ in Eq~\ref{eq1}). Finally, according to the demand of the network, the feature map resolution is pooled to acquire the output of the corresponding resolution (Denoted as $Out_{s}^{i},i\in(0,2)$ and $Out_{l}^{j},j\in(0,1)$ in Eq~\ref{eq2}).

\vspace{-0.5em}
\begin{small}
\begin{equation}
\centering
\label{eq1}
\left\{\begin{matrix}
SFm_{s}=\left ( P\left ( fm^{0} \right )+P\left ( fm^{1} \right ) \right ),   \\
SFm_{l}=\left ( P\left ( fm^{2} \right )+P\left ( fm^{3} \right ) \right ) ,
\end{matrix}\right.
\end{equation}
\end{small}

\vspace{-0.3em}
\begin{small}
\begin{equation}
\centering
\label{eq2}
\left\{\begin{matrix}
Out_{s}^{k} =(c^{*} (p^{*}_{2^k}(SFm_{s}))&k\in(0,2),\\ 
Out_{l}^{j} =(c^{*} (p^{*}_{2^j}(SFm_{l}))&j\in(0,1) ,
\end{matrix}\right.
\end{equation}
\end{small}

\noindent where $c^{*}$ and $p^{*}$ indicate convolution layer and pooling layer, respectively. $2^{k}$ and $2^{j}$ indicate the size of the dimensions in the pooling layer. Eq~\ref{eq1} expresses the calculation process of the feature map in the two parallel feature aggregation heads. More specifically, the multi-head module obtains two aggregated feature maps by measuring the output feature maps in four different dimensions, interpolating and concatenating the feature maps in adjacent dimensions to the same size. Eq~\ref{eq2} illustrates the process of differentiation from the semantic information-rich $SFm$ into the corresponding output feature maps. The heads are pooled several times to get the corresponding size. 

In particular, in order to achieve targeted recovery of the network's focus on multi-resolution feature maps, we use parallel two-headed modules for high-resolution and low-resolution features separately. Compared with the single-head module (HRFPN), although the multi-head module increases the calculation volume, it can attenuate the effect on focus inequality due to the difference in the ratio of breast mass images. Moreover, the approach can shorten the path of information flow insertion to the final resolution and prevent the feature information from being scattered again. On the feature maps of neighboring dimensions in the same group, the higher-level semantic information is calculated by focus reorganization with neighboring features using up-sampling to retain and improve the influence of higher-level semantic information on $SFm$. MHFPN has correspondingly improved the detection effect both on small targets and on large targets, and it works better on small targets.
\vspace*{-2.0ex}
\section{EXPERIMENTAL Studies}
\vspace*{-1ex}
\label{sec:experiment}

\subsection{Experimental Environment and Dataset}
\vspace{-0.3em}
\label{subsec:DATA IN Experiment}
All experiments were run on a server with 2 Nvidia GeForce 2080 GPUs with 8192M RAM per card. The server has an Intel(R) Xeon(R) Silver 4110 CPU with 2.10GHz and 16G RAM. For the software configuration, all code was implemented in Python based on the mmdetection framework. 
\begin{small}
\begin{table}[!t]
\centering
\vspace{-0.5em}  
\caption{Detection results on three breast masses datasets.}
\label{table:5}
\vspace{0.3em}
\begin{tabular}{cccc}
\hline
Dataset      & Method      & AP@20 & TPR@20 \\ \hline
INbreast    & Faster RCNN  & 0.725 & 0.933  \\
            & \textbf{MBMDNet}         & 0.806 & 0.962  \\
BCS-DBT~\cite{buda2021data}        & Faster RCNN  & 0.350 & 0.881  \\
            & \textbf{MBMDNet}         & 0.412 & 0.878  \\
MIAS~\cite{suckling1994mammographic}       & Faster RCNN  & 0.312 & 0.535  \\
            & \textbf{MBMDNet}         & 0.396 & 0.619  \\ \hline
\end{tabular}
\vspace{-1.5em}    
\end{table}
\end{small}

\begin{small}
\begin{table}[!t]
\centering
\vspace{-0.5em}
\caption{ Compare to advanced detection models on INbreast.}\vspace{0.3em}
\label{table:4}
\begin{tabular}{ccc}
\hline
 Methods                                        & TPR@20  & FPPI        \\ \hline
 Amit et al.\cite{amit2015automatic}            & 0.87    & 1.423       \\
 Wu et al.\cite{wu2018automatic}                & 0.88    & 0.750        \\
 Akselrod-Ballin et al.~\cite{akselrod2017deep} & 0.93    & 0.560       \\
 BMassDNet~\cite{cao2021breast}      & 0.93    & \textbf{0.495}  \\
 \textbf{MBMDNet}                                   & \textbf{0.96}    & 0.636 \\  \hline
\end{tabular}
\vspace{-1.5em}    
\end{table}
\end{small}

\begin{small}
\begin{table*}[!t]
\centering
\vspace{-2em}  
\caption{Results of ablation experiments on the INbreast dataset}
\vspace{0.5em}
\label{table:2}
\begin{tabular}{ccccccc}
\hline
 Net                              & AP@50           & AP@50S         &AP@50L          & TPR@50          & FLOPS    & Params  \\ \hline
 FPN                              & 0.5804 (0.0730)  & 0.1470 (0.0210)  & 0.6020 (0.0540) &0.7850 (0.0585)  & 216.30G  & 41.13M  \\
 HRFPN    &0.5701 (0.0929) &0.0840 (0.0801) &0.5530 (0.0290) &0.8006 (0.0982) &274.24G &41.72M \\
 PANet         & 0.5974 (0.1070)  & -               & 0.5560 (0.0390) &0.7964 (0.1505)  & 242.33G  & 44.67M \\
 MHFPN   & \textbf{0.6462 (0.0934)}  & \textbf{0.3260 (0.0740)} & \textbf{0.6345 (0.0295)} &\textbf{0.8390 (0.0915)}  & 291.36G & 44.80M \\ \hline
\end{tabular}
\vspace{-1em}    
\end{table*}
\end{small}
The main dataset used for the experiment is the INbreast dataset \footnote{\url{http://medicalresearch.inescporto.pt}}, which is more commonly used for breast mass detection. The dataset has 107 images containing 116 masses. Each image had an average of 1.1 masses, with the smallest mass being 15 $mm^{2}$ and the largest mass measuring 3689 $mm^{2}$. Meanwhile, in order to test the performance of the network on other different types of datasets, the earlier MIAS dataset \footnote{\url{http://peipa.essex.ac.uk/info/mias.html}} and the latest BCS-DBT dataset \footnote{\url{https://wiki.cancerimagingarchive.net/}} are also chosen. For the accuracy of the experimental data, every dataset is divided equally into five parts, of which three parts are the training set, one part is the validation set, and one part is the test set. Five replicate experiments are conducted by randomizing permutations of the five data parts. The data are processed slightly differently depending on the experimental objectives. Please refer to the corresponding experimental notes for detailed processing.
\vspace{-1.0em}
\subsection{Main Results}
\vspace{-0.5em}
\label{subsec:INbreast test}
In the paper, a variety of classical target detection networks are selected for training and testing on the INbreast dataset. The best-performing network is selected by comparing the common parameter metrics~\cite{zhang2022high} such as $AP@50$ and $AP@75$. In the test metrics, $AP$ represents the average accuracy rate. $TPR$ is the True Positive Rate (i.e., Recall). $@x$ denotes the calculation of the relevant evaluation metric under the restriction with the $IOU > x$. $FLOPS$ and $Params$ represent the computational and parametric quantities of the model, respectively. As can be seen from Table~\ref{table:1} (The data in the table are the mean (standard deviation)), compared to other classical networks, Faster RCNN achieves 58.04\% and 31.60\% in $AP@50$ and $AP@75$, respectively. Thus MBMDnet is based on Faster RCNN and achieves 64.62\% in the $AP@50$ and 83.90\% in the $TPR@50$.

Fig~\ref{fig:MHFPN compare} visualizes the inference of different networks on the INbreast dataset. In example 1, there are adjacent closer regions with larger size masses and smaller size masses. From the inference results, it can be observed that MBMDNet is more effective in detecting such targets with larger size differences. In examples 2 and 3, the detected breast mass is the only mass in the entire image and is small in size. Compared with others, the inference results of MBMDNet fit the ground truth box more closely.

To further test the generalizability of the network under different quality datasets, MBMDnet and Faster RCNN are evaluated on both MIAS dataset~\cite{suckling1994mammographic} and BCS-DBT dataset~\cite{buda2021data}. The above two datasets are treated in the same way as the INbreast dataset. Due to the early production of the MIAS dataset, the quality and annotation of the images differed significantly from the current approach. The dataset showed a large amount of blurred noise, which seriously affected the correct rate of breast mass detection. The BCS-DBT dataset is a recently proposed large dataset for breast lesions. The dataset file has more image data and less annotation data. The images are darker in general and there is interference from non-breast objects, which is more consistent with the real clinical situation.

From Table~\ref{table:5}, it can be seen that there is a large difference in performance on different quality data sets. But compared to the Faster RCNN network, our approach still has a 6\%-8\% improvement in $AP@20$. In the $TPR@20$ metric, our method also shows an improvement of about 3.7\% without fine-tuning additional procedures.

To further compare the performance of MBMDNet, we chose the advanced detection models with the INbreast dataset of 107 images as the dataset for comparison. $FPPI$ represents False Positive per image. As can be seen in Table~\ref{table:4} (The data in the table are the maximum values from five experiments), Although our method does not perform well in the $FPPI$, it reaches 0.96 in the $TPR@20$ metric. In medical computer-aided systems, a missed test is more harmful to the judgment of a medical condition than a false test. And in the network training, we use Free-response Receiver Operating Characteristic Curves (FROC) to select the best TPR performance. Therefore, our method is not the best model in terms of performance on the $FPPI$ metric.
\vspace{-1.0em}
\subsection{Ablation Study}
\label{subsubsec: Ablation Study}
\vspace{-0.4em}
To test the effectiveness of each part of the MHFPN module, we performed an ablation analysis of the information flow channels during the feature fusion phase of the target detection network. Meanwhile, in order to target test the degree to the network's equalization of focus to target boxes with different size ratios in the feature fusion, the accuracy of different sizes of target boxes ($AP@S$ and $AP@L$) is added to the evaluation metrics of the ablation experiment. According to the format of the COCO data, the size of the breast mass is divided into large, medium and small boxes, and then the performance metrics are tested. $AP@50S$ indicates the detection accuracy of the small box under the condition that $IOU > 0.5$. $AP@50L$ is the detection accuracy of a large box under the same constraints.

\begin{small}
\begin{figure*}[!t]
\vspace{-2em}
\centering
\includegraphics[width=140mm]{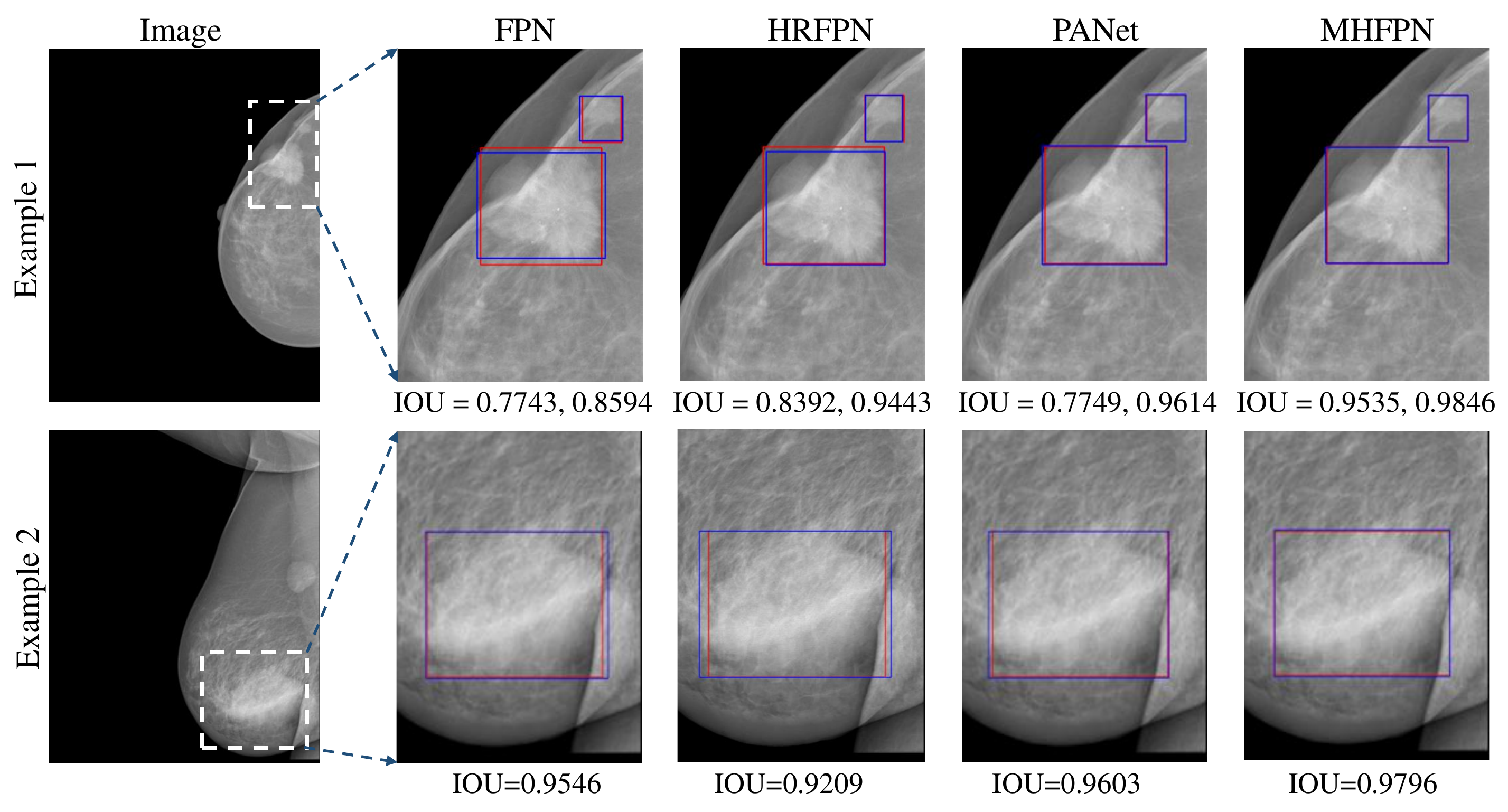}
\vspace{-1em}
\caption{Visualized results of ablation experiments on INbreast, and blue (resp., red) boxes are detection results (resp., ground truths).}
\label{fig:MHFPN ABstudy}
\vspace{-1.5em}
\end{figure*}
\end{small}

According to Table~\ref{table:2}, compared to the FPN network, both PANET and HRFPN had reduced detection accuracy for targets with a smaller ratio of x-ray breast masses. Thus, it is demonstrated that the detection of small targets in X-ray images by simply adding information flow paths or information aggregation does not perform well. Although MHFPN increases the model computation, the overall detection accuracy is improved by 6\%$AP@50$. In particular, it improves the detection accuracy of small targets by 17.9\%, thus proving that the parallel multi-head network structure can alleviate the problem of unbalanced focus of target boxes. From the visualization in Fig~\ref{fig:MHFPN ABstudy}, it can be observed that when there are two adjacent target boxes of different sizes (such as Example 1), the other three networks have improved accuracy compared to the FPN network. Specifically, the MHFPN network has the highest IOU accuracy in terms of specific detection performance. Additionally, for larger target boxes (such as Example 2), MHFPN also achieves the best results.

\vspace*{-1ex}
\section{CONCLUSIONS And Future Works}
\vspace*{-0.5ex}
\label{sec:conclusion}

In the paper, we designed a general multi-head feature pyramid module (MHFPN) to address the problem of unequal focus to the target frame by multi-resolution feature maps in the feature fusion. Based on MHFPN, the detection network (MBMDNet) was designed for breast mass detection. Experiments showed that MBMDNet greatly outperformed the SOTA detection baselines on multiple breast X-ray datasets. In particular, the overall performance of the network was improved more in the detection accuracy of targets with a smaller image. However, this work can be further improved, to address the problem of lacking sufficient labels, an interesting future research is to investigate semi-supervised approaches~\cite{ZHANG-MIA2022,w-Net2022,xu2019semi,DRLHomoPre2022} to utilize the unlabeled data in the training process of the model.

\vspace*{-2ex}
\section{ACKNOWLEDGMENTS}
\vspace*{-1.5ex}
This work was supported by the National Natural Science Foundation of China under the grants 62276089, 61906063 and 62102265, by the Natural Science Foundation of Hebei Province, China, under the grant F2021202064, by the "100 Talents Plan" of Hebei Province, China, under the grant E2019050017, by the Open Research Fund from Guangdong Laboratory of Artificial Intelligence and Digital Economy (SZ) under the grant GML-KF-22-29, by the Natural Science Foundation of Guangdong Province of China under the grant 2022A1515011474.
\clearpage
\begin{spacing}{1.0}
  \bibliographystyle{IEEEbib}
  \small
  \bibliography{IEEE.bib}

\begin{thebibliography}{10}

\bibitem{bray2018global}
Freddie Bray, Jacques Ferlay, Isabelle Soerjomataram, Rebecca~L Siegel,
  Lindsey~A Torre, and Ahmedin Jemal,
\newblock ``Global cancer statistics 2018: Globocan estimates of incidence and
  mortality worldwide for 36 cancers in 185 countries,''
\newblock {\em CA: A Cancer Journal for Clinicians}, vol. 68, no. 6, pp.
  394--424, 2018.

\bibitem{yu2022systematic}
Xiang Yu, Qinghua Zhou, Shuihua Wang, and Yu-Dong Zhang,
\newblock ``A systematic survey of deep learning in breast cancer,''
\newblock {\em Proceedings of International Journal of Intelligent Systems},
  vol. 37, no. 1, pp. 152--216, 2022.

\bibitem{gastounioti2022artificial}
Aimilia Gastounioti, Shyam Desai, Vinayak~S Ahluwalia, Emily~F Conant, and
  Despina Kontos,
\newblock ``Artificial intelligence in mammographic phenotyping of breast
  cancer risk: a narrative review,''
\newblock {\em Breast Cancer Research}, vol. 24, no. 1, pp. 1--12, 2022.

\bibitem{zhang2022anchor}
Linlin Zhang, Yanfeng Li, Houjin Chen, Wen Wu, Kuan Chen, and Shaokang Wang,
\newblock ``Anchor-free {YOLOv3} for mass detection in mammogram,''
\newblock {\em Expert Systems with Applications}, vol. 191, pp. 116273, 2022.

\bibitem{ibrokhimov2022two}
Bunyodbek Ibrokhimov and Justin-Youngwook Kang,
\newblock ``Two-stage deep learning method for breast cancer detection using
  high-resolution mammogram images,''
\newblock {\em Applied Sciences}, vol. 12, no. 9, pp. 4616, 2022.

\bibitem{su2022yolo}
Yongye Su, Qian Liu, Wentao Xie, and Pingzhao Hu,
\newblock ``{YOLO-LOGO}: A transformer-based yolo segmentation model for breast
  mass detection and segmentation in digital mammograms,''
\newblock {\em Computer Methods and Programs in Biomedicine}, p. 106903, 2022.

\bibitem{PAC-net}
Zhenghua Xu, Tianrun Li, Yunxin Liu, Yuefu Zhan, Junyang Chen, and Thomas
  Lukasiewicz,
\newblock ``{PAC-Net}: Multi-pathway {FPN} with position attention guided
  connections and vertex distance {IoU} for {3D} medical image detection,''
\newblock {\em Frontiers in Bioengineering and Biotechnology}, vol. 11, pp.
  1049555, 2023.

\bibitem{wu2018automatic}
Yifan Wu, Weifeng Shi, Lei Cui, Hongyu Wang, Qirong Bu, and Jun Feng,
\newblock ``Automatic mass detection from mammograms with region-based
  convolutional neural network,''
\newblock in {\em Proceedings of Chinese Conference on Image and Graphics
  Technologies}, 2018, pp. 442--450.

\bibitem{cao2021breast}
Haichao Cao, Shiliang Pu, Wenming Tan, and Junyan Tong,
\newblock ``Breast mass detection in digital mammography based on anchor-free
  architecture,''
\newblock {\em Computer Methods and Programs in Biomedicine}, vol. 205, pp.
  106033, 2021.

\bibitem{moreira2012inbreast}
In{\^e}s~C Moreira, Igor Amaral, In{\^e}s Domingues, Ant{\'o}nio Cardoso,
  Maria~Joao Cardoso, and Jaime~S Cardoso,
\newblock ``I{N}breast: toward a full-field digital mammographic database,''
\newblock {\em Academic Radiology}, vol. 19, no. 2, pp. 236--248, 2012.

\bibitem{redmon2018yolov3}
Joseph Redmon and Ali Farhadi,
\newblock ``{YOLOv3}: An incremental improvement,''
\newblock {\em arXiv preprint arXiv:1804.02767}, 2018.

\bibitem{lin2017focal}
Tsung-Yi Lin, Priya Goyal, Ross Girshick, Kaiming He, and Piotr Doll{\'a}r,
\newblock ``Focal loss for dense object detection,''
\newblock in {\em Proceedings of the IEEE International Conference on Computer
  Vision}, 2017, pp. 2980--2988.

\bibitem{liu2016ssd}
Wei Liu, Dragomir Anguelov, Dumitru Erhan, Christian Szegedy, Scott Reed,
  Cheng-Yang Fu, and Alexander~C Berg,
\newblock ``Ssd: Single shot multibox detector,''
\newblock in {\em Proceedings of European Conference on Computer Vision}, 2016,
  pp. 21--37.

\bibitem{wang2020fcos}
Ning Wang, Yang Gao, Hao Chen, Peng Wang, Zhi Tian, Chunhua Shen, and Yanning
  Zhang,
\newblock ``Nas-fcos: Fast neural architecture search for object detection,''
\newblock in {\em Proceedings of the IEEE/CVF Conference on Computer Vision and
  Pattern Recognition}, 2020, pp. 11943--11951.

\bibitem{cai2018cascade}
Zhaowei Cai and Nuno Vasconcelos,
\newblock ``Cascade r-cnn: Delving into high quality object detection,''
\newblock in {\em Proceedings of the IEEE Conference on Computer Vision and
  Pattern Recognition}, 2018, pp. 6154--6162.

\bibitem{ren2015faster}
Shaoqing Ren, Kaiming He, Ross Girshick, and Jian Sun,
\newblock ``Faster r-cnn: Towards real-time object detection with region
  proposal networks,''
\newblock {\em Advances in Neural Information Processing Systems}, vol. 28,
  2015.

\bibitem{liu2018path}
Shu Liu, Lu~Qi, Haifang Qin, Jianping Shi, and Jiaya Jia,
\newblock ``Path aggregation network for instance segmentation,''
\newblock in {\em Proceedings of the IEEE Conference on Computer Vision and
  Pattern Recognition}, 2018, pp. 8759--8768.

\bibitem{sun2019high}
Ke~Sun, Yang Zhao, Borui Jiang, Tianheng Cheng, Bin Xiao, Dong Liu, Yadong Mu,
  Xinggang Wang, Wenyu Liu, and Jingdong Wang,
\newblock ``High-resolution representations for labeling pixels and regions,''
\newblock {\em arXiv preprint arXiv:1904.04514}, 2019.

\bibitem{buda2021data}
Mateusz Buda, Ashirbani Saha, Ruth Walsh, Sujata Ghate, Nianyi Li, Albert
  {\'S}wiecicki, Joseph~Y Lo, and Maciej~A Mazurowski,
\newblock ``A data set and deep learning algorithm for the detection of masses
  and architectural distortions in digital breast tomosynthesis images,''
\newblock {\em JAMA Network Open}, vol. 4, no. 8, pp. e2119100--e2119100, 2021.

\bibitem{suckling1994mammographic}
P~Suckling~J,
\newblock ``The mammographic image analysis society digital mammogram
  database,''
\newblock {\em Digital Mammo}, pp. 375--386, 1994.

\bibitem{amit2015automatic}
Guy Amit, Sharbell Hashoul, Pavel Kisilev, Boaz Ophir, Eugene Walach, and Aviad
  Zlotnick,
\newblock ``Automatic dual-view mass detection in full-field digital
  mammograms,''
\newblock in {\em Proceedings of International Conference on Medical Image
  Computing and Computer-Assisted Intervention}, 2015, pp. 44--52.

\bibitem{akselrod2017deep}
Ayelet Akselrod-Ballin, Leonid Karlinsky, Alon Hazan, Ran Bakalo, Ami~Ben
  Horesh, Yoel Shoshan, and Ella Barkan,
\newblock ``Deep learning for automatic detection of abnormal findings in
  breast mammography,''
\newblock in {\em Proceedings of DLMIA workshops}, pp. 321--329. 2017.

\bibitem{zhang2022high}
Hexiang Zhang, Xiaofang Yang, Ziyu Hu, Ruoxin Hao, Zehang Gao, and Jianhao
  Wang,
\newblock ``High-density pedestrian detection algorithm based on deep
  information fusion,''
\newblock {\em Applied Intelligence}, pp. 1--13, 2022.

\bibitem{ZHANG-MIA2022}
Shuo Zhang, Jiaojiao Zhang, Biao Tian, Thomas Lukasiewicz, and Zhenghua Xu,
\newblock ``Multi-modal contrastive mutual learning and pseudo-label
  re-learning for semi-supervised medical image segmentation,''
\newblock {\em Medical Image Analysis}, vol. 83, pp. 102656, 2023.

\bibitem{w-Net2022}
Zhenghua Xu, Shijie Liu, Di~Yuan, Lei Wang, Junyang Chen, Thomas Lukasiewicz,
  Zhigang Fu, and Rui Zhang,
\newblock ``$\omega$-net: Dual supervised medical image segmentation with
  multi-dimensional self-attention and diversely-connected multi-scale
  convolution,''
\newblock {\em Neurocomputing}, vol. 500, pp. 177--190, 2022.

\bibitem{xu2019semi}
Zhenghua Xu, Chang Qi, and Guizhi Xu,
\newblock ``Semi-supervised attention-guided {CycleGAN} for data augmentation
  on medical images,''
\newblock in {\em Proceedings of the IEEE International Conference on
  Bioinformatics and Biomedicine}, 2019, pp. 563--568.

\bibitem{DRLHomoPre2022}
Di~Yuan, Yunxin Liu, Zhenghua Xu, Yuefu Zhan, Junyang Chen, and Thomas
  Lukasiewicz,
\newblock ``Painless and accurate medical image analysis using deep
  reinforcement learning with task-oriented homogenized automatic
  pre-processing,''
\newblock {\em Computers in Biology and Medicine}, vol. 153, pp. 106487, 2023.

\end{thebibliography}
\end{spacing}
\end{document}